\documentclass[conference]{IEEEtran}
\IEEEoverridecommandlockouts
\usepackage{cite}
\usepackage{amsmath,amssymb,amsfonts}
\usepackage{algorithmic}
\usepackage{graphicx}
\usepackage{textcomp}
\usepackage{xcolor}
\usepackage{hyperref}
\usepackage{diagbox} %JON
\usepackage{comment} %JON
\usepackage{multicol} %JON
\usepackage{tikz}
\usepackage[absolute,overlay]{textpos}
\usepackage{blindtext}
\usepackage{relsize}
\usepackage{hyperref}

\tikzset{fontscale/.style={font=\relsize{#1}}}

\def\BibTeX{{\rm B\kern-.05em{\sc i\kern-.025em b}\kern-.08em
    T\kern-.1667em\lower.7ex\hbox{E}\kern-.125emX}}
\begin{document}

\title{Rapid Deployment of Domain-specific Hyperspectral Image Processors with Application to Autonomous Driving*
\thanks{*This work was partially supported by the Basque Government under grants PRE\_2022\_2\_0210 and KK-2023/00090, by the Spanish Ministry of Science and Innovation under grant PID2020-115375RB-I00 and by the University of the Basque Country (UPV-EHU) under grant GIU21/007.}
}

\author{\IEEEauthorblockN{1\textsuperscript{st} Jon Gutiérrez-Zaballa}
\IEEEauthorblockA{\textit{Department of Electronics Technology} \\
\textit{University of the Basque Country}\\
Bilbao, Spain \\
0000-0002-6633-4148}
\and
\IEEEauthorblockN{2\textsuperscript{nd} Koldo Basterretxea}
\IEEEauthorblockA{\textit{Department of Electronics Technology} \\
\textit{University of the Basque Country}\\
Bilbao, Spain \\
0000-0002-5934-4735}
\and
\IEEEauthorblockN{3\textsuperscript{rd} Javier Echanobe}
\IEEEauthorblockA{\textit{Department of Electricity and Electronics} \\
\textit{University of the Basque Country}\\
Leioa, Spain \\
0000-0002-1064-2555}
\and
\IEEEauthorblockN{4\textsuperscript{th} Óscar Mata-Carballeira}
\IEEEauthorblockA{\textit{Department of Electronics Technology} \\
\textit{University of the Basque Country}\\
Bilbao, Spain \\
0000-0002-1468-6280}
\and
\IEEEauthorblockN{5\textsuperscript{th} M. Victoria Martínez}
\IEEEauthorblockA{\textit{Department of Electricity and Electronics} \\
\textit{University of the Basque Country}\\
Leioa, Spain \\
}}

\maketitle
\begin{textblock*}{21cm}(1.5cm,26cm)
  \begin{tikzpicture}
    \draw (0,0) rectangle (18,0.5); 
    \end{tikzpicture}
\end{textblock*} 

\begin{textblock*}{21cm}(0cm,26cm)
  \begin{tikzpicture}
    \node (center) {c};
    \path (center)+(10.5,4) node [fontscale=-1] (name) {\copyright 2024 IEEE. Final published version of the article can be found at \href{https://ieeexplore.ieee.org/document/10382745}{10.1109/ICECS58634.2023.10382745}.};
    \end{tikzpicture}
\end{textblock*}

\begin{abstract}
The article discusses the use of low cost System-On-Module (SOM) platforms for the implementation of efficient hyperspectral imaging (HSI) processors for application in autonomous driving.
The work addresses the challenges of shaping and deploying multiple layer fully convolutional networks (FCN) for low-latency, on-board image semantic segmentation using resource- and power-constrained processing devices.
The paper describes in detail the steps followed to redesign and customize a successfully trained HSI segmentation lightweight FCN that was previously tested on a high-end heterogeneous multiprocessing system-on-chip (MPSoC) to accommodate it to the constraints imposed by a low-cost SOM.
This SOM features a lower-end but much cheaper MPSoC suitable for the deployment of automatic driving systems (ADS).
In particular the article reports the data- and hardware-specific quantization techniques utilized to fit the FCN into a commercial fixed-point programmable AI coprocessor IP, and proposes a full customized post-training quantization scheme to reduce computation and storage costs without compromising segmentation accuracy.
\end{abstract}


\section{Introduction}
The advent of small-size, snapshot-type hyperspectral cameras has enabled the widespread use of hyperspectral imaging (HSI) in new application domains \cite{fricker2019convolutional, taghizadeh2011comparison, seidlitz2022robust}.
HSI provides valuable information about how materials reflect different light wavelengths (spectral reflection), which can be used to detect and segment surfaces and objects in a scene \cite{weikl2022potentials}.
Advanced driver assistance systems (ADAS) and autonomous driving systems (ADS) are particularly promising targets for this technology.
However, the successful deployment of HSI-aided ADS requires the production of high-performance processing systems that are cost-effective and have low power consumption.
In this context, adaptive System on Modules (SOMs) based on multiprocessing system-on-chip (MPSoC) devices have emerged as an interesting technological choice for implementing high-performance, AI-enabled ADS with reduced development time and cost.

In this study, we present the process of adapting and recustomizing a previously developed AI-enabled HSI segmentation system for a mosaic-filter snapshot hyperspectral camera.
We specifically focus on the steps taken to achieve a successful 8-bit quantized model that can be efficiently processed on these devices while meeting the latency requirements imposed by this application.
Additionally, we provide performance metrics for the image segmentation system running on an AMD-Xilinx Kria K26 SOM, consuming 7.6W.

\section{FCN model development}
In the research presented in this article, we built upon our previous findings: a tiny FCN model that required the composition of multiple small-size image patches, which is described in \cite{gutierrez2023chip}, and a deeper, more capable FCN that is presented in \cite{civts2023}.
In order to maximize processing performance, we have redesigned the larger FCN reducing the model depth from 5 to 4, allowing the processing of a complete hyperspectral cube in a single pass, thus reducing 4x the number of parameters without significant degradation in accuracy (Table \ref{tab:compCompFloat}).
By processing the entire image at once, we eliminate the time and memory overhead associated with reconstructing the full image from overlapping patches.
Moreover, processing the whole image allows for a larger context to be captured for the extraction of spatial features, although it requires increasing the depth of the FCN.
As a general rule, in encoder-decoder architectures, increasing the depth-level by one implies reducing the spatial size by a factor of 4.
The increase in model size is not solely due to larger image sizes, but also to the utilization of the recently published extended version of the HSI-Drive dataset, HSI-Drive v2.0.
This extended dataset has enabled the training of a more robust and accurate model, albeit at the cost of more trainable parameters.
For a comprehensive review of the dataset, we refer the reader to the website \url{https://ipaccess.ehu.eus/HSI-Drive/}, where the this dataset is available upon request.

\begin{table}[t]
\centering
\caption{Computational complexity and achieved IoU for the 32-bit floating point FCNs.}
\label{tab:compCompFloat}
\begin{tabular}{|c|c|c|}
\cline{1-3}
\textbf{Depth-level} & \textbf{5} & \textbf{4} \\ \hline
\multicolumn{1}{|c|}{Num. of K params.} & 31 135.062 & 7 778.374 \\ \hline
\multicolumn{1}{|c|}{Num. of trainable K params.} & 31 113.030 & 7 772.486 \\ \hline
\multicolumn{1}{|c|}{Num. of GFLOPS} & 34.596 & 31.797 \\ \hline
\multicolumn{1}{|c|}{Global IoU} & 94.64 & 92.28 \\ \hline
\end{tabular}
\end{table}

The architecture of the lightweight FCN used in this study is a modification of the model depicted in Fig. 6 in \cite{gutierrez2023chip}.
It consists of 32 filters in the first convolutional block, 3x3 convolution kernels and 4 levels of depth.
Consequently, the input image size is restricted to be a multiple of $2^4$, which results in 208x400 pixels for this model.
This involves framing the image by cropping 8/9 pixels from the edges.
In this article, we adopt a previously designed 5-class classification system as the reference model for implementation, which was designed to segment the images into Tarmac, Road Marks, Vegetation, Sky and "Others" classes.
This segmentation is useful to differentiate the Road (Tarmac and Road Marks) from the background (Vegetation and Sky typically) and from unknown objects and obstacles such as vehicles, cyclist or pedestrians or informative still objects such as road signs, traffic lights and information panels \cite{gutierrez2023chip}.

The FCNs have been codified using Keras/Tensorflow2 frameworks and have been trained on a NVIDIA GFORCE RTX-3090 with 24GB of memory.
The minimum weighted cross-entropy loss (normalized inverse frequency weighting) has been obtained with an Adam optimizer with an initial learning rate of 0.001, gradient decay factor of 0.9, squared gradient decay factor of 0.999, 200 epochs, data shuffling at each epoch and a train batch-size of 23 images and a validation batch-size of 49 images.
To mitigate the impact of random initialization, each training has been repeated 3 times.
Besides, the dataset has been homogeneously structured for a 5-fold cross-validation process taking into account the features of the images in the dataset, i.e. daytime, climatology, season and road type.

\section{Model quantization}
\begin{figure}[b]
\centerline{\includegraphics[width=0.45\textwidth]{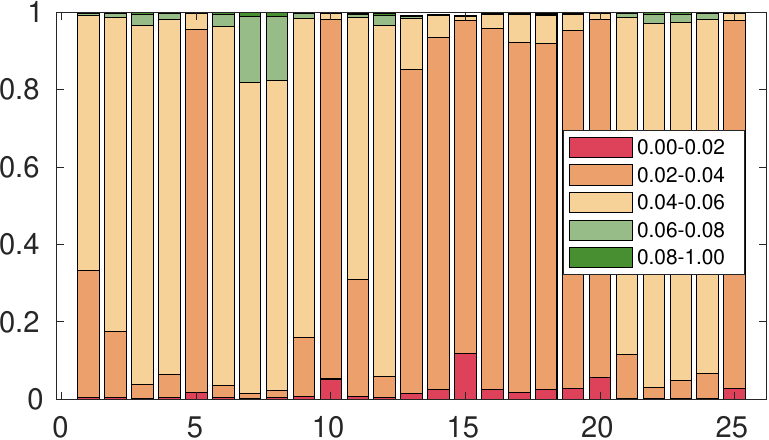}}
\caption{Input data distribution for the 25 spectral channels.}
\label{fig:dataDistribution}
\end{figure}

Quantizing the FCN is a necessary step to deploy the model on the fixed-point arithmetic AMD-Xilinx Deep Learning Processor Unit (DPU) \cite{dpu} AI coprocessor.
However, achieving a successful post-training quantization that keeps segmentation accuracy requires carefully analyzing the model parameters and signal ranges, and potentially applying tailored adaptations.
In fact, applying a range-preserving quantization method directly to this model resulted in poor performance, with a decrease in IoU of over 45\%.
A thorough analysis of the signal ranges reveals that the normalization method applied to the reflectance values of the cubes in the preprocessing stage (see \cite{civts2023}) leads to data accumulation around the value 0.04, which corresponds to the inverse of the number of spectral channels, as shown in Fig. \ref{fig:dataDistribution}.
As depicted, the [0, 0.08] range contains 99.7175\% of all pixels.

As a result, we first performed an adaptive clipping of the normalized reflectance values based on the data distribution in each spectral channel.
The obtained clipping values ranged from 0.0711 to 0.1495, ensuring that 99.95\% of the data are accurately represented.
Applying this method allows for saving 3 integer bits to increase the resolution of the fixed point number representation in the quantization process.
After retraining the model using these savings, we verified that the overall performance of the floating-point clipped range network was practically unaffected, with IoU index variations of +0.086 for Road, -1.146 for Road Marks, +1.014 for Vegetation, +1.574 for Sky and +0.182 for Others.
Furthermore, and more importantly, as explained below, the obtained model proved to be robust to Min-Max quantization to 8-bit precision.

Secondly, since some of the quantization techniques and segmentation recovery methods that aid to produce accurate quantized models require the activation functions of the convolutional layers to be piecewise linear, all activation functions in the original FCN model were set to ReLU units.
This is particularly important for techniques like cross layer equalization (CLE), which tries to minimize the difference in magnitude of the elements in the same tensor without the need to use per-channel quantization \cite{nagel2019data}.
Similarly, bias absorption is a segmentation recovery method which intends to decrease the differences in the dynamic ranges of activations, especially after applying CLE.
In that case, a special requirement is to use the ReLU activation function \cite{nagel2021white}.

With all this in mind, we performed a customized quantization pipeline of the lightweight FCN model (inputs, weights, bias and activations) using AMD-Xilinx Vitis AI 3.0 tool.
This tool imposes some restrictions.
First, the quantization scheme must be homogeneous and uniform, i.e. all layers have the same bit-width and every interval has assigned the same quantization level.
Secondly, the scale factors of the quantization are restricted to be powers of two.
The quantization scheme, as depicted in \ref{equ:quantizationScheme}, can be asymmetric or symmetric (zero-point is restricted to 0):

\begin{equation}
\textbf{x} \approx \hat{\textbf{x}} = s_{\textbf{x}}(\textbf{x}_{int} - z_{\textbf{x}}),
\label{equ:quantizationScheme}
\end{equation}

where $\hat{\textbf{x}}$ is the approximation of the real-valued $\textbf{x}$, $s_{\textbf{x}}$ is the scale factor, $\textbf{x}_{int}$ is the unsigned integer mapping of the real-valued $\textbf{x}$ and $z_{\textbf{x}}$ is the zero-point.
We applied symmetric quantization for inputs (spectral cubes are normalized to [-1.0, +1.0) for model training) and weights (weight distributions are symmetric along the model layers) while we chose asymmetric quantization for activations to save the sign bit in the data representation (ReLUs produce one-tailed distributions).
The combination of symmetric weight quantization with asymmetric activation quantization produces a good trade-off between precision and processing performance.
If a general asymmetric weight ($\textbf{W}$) and activation representation is applied, the second term in \ref{equ:symmetricAsymmetricQuantization} would have to be computed on-the-fly during inference as it depends on the value of input data $\textbf{x}$.
On the contrary, in the selected quantization scheme the terms $z_{\textbf{w}}$ are $\textbf{0}$, so the second and fourth terms in \ref{equ:symmetricAsymmetricQuantization} become both null.
As the third term only depends on the scale factor, the offset and the weight values, it can be pre-computed and added to the bias term of the layer with no extra processing overhead \cite{nagel2021white}:

\begin{equation}
\begin{split}
\hat{\textbf{W}}\hat{\textbf{x}} & = s_{\textbf{w}}(\textbf{W}_{int} - z_{\textbf{w}}) s_{\textbf{x}}(\textbf{x}_{int} - z_{\textbf{x}}) \\ & = s_{\textbf{w}}s_{\textbf{x}}\textbf{W}_{int}\textbf{x}_{int} - s_{\textbf{w}}z_{\textbf{w}}s_{\textbf{x}}\textbf{x}_{int} - s_{\textbf{w}}s_{\textbf{x}}z_{\textbf{x}}\textbf{W}_{int} + s_{\textbf{w}}z_{\textbf{w}}s_{\textbf{x}}z_{\textbf{x}}
\end{split}
\label{equ:symmetricAsymmetricQuantization}
\end{equation}

Regarding quantization granularity, we opted for a per-tensor quantization procedure since, as explained in \cite{nagel2021white}, hardware implementation becomes more complex (especially for activations) if per-channel quantization is applied and it is recommended only when performance improvement is required.
In relation to the quantization methods, and since we performed a previous signal range clipping procedure to maximize precision, we applied range-preserving Min-Max quantization both for inputs and biases, whose correct quantization is essential not to generate an overall bias that would compromise accuracy, \cite{jacob2018quantization}.
As for weights and activations, we opted for a Min-MSE quantization to minimize the distance between original and quantized tensors and neglect large outliers.
Finally, besides the quantization of parameters and activation functions, cross layer equalization and Batch Normalization folding \cite{ioffe2015batch} were also applied to reduce the number of operations in the inference phase.
Table \ref{tab:compCompQuant} shows the computational complexity of the obtained 8-bit quantized FCN.

\begin{table}[h!]
\centering
\caption{Computational complexity of the 8-bit quantized FCN. OPS represent the total number of Operations. Model size represents the occupation of weights and biases.}
\label{tab:compCompQuant}
\begin{tabular}{|c|c|}
\cline{1-2}
\multicolumn{1}{|c|}{Num. of K params.} & 7 766.886 \\ \hline
\multicolumn{1}{|c|}{Num. of trainable K params.} & 7 766.598 \\ \hline
\multicolumn{1}{|c|}{Num. of OPS} & 31.761 \\ \hline
\multicolumn{1}{|c|}{Model size in MB} & 7.407 \\ \hline
\end{tabular}
\end{table}

\section{Network deployment and testing}
ADAS/ADS applications need to adhere to demanding constraints in terms of energy consumption, cost, and processing latency.
Furthermore, the rapid evolution of neural network models requires processing hardware to be adaptable in order to avoid the obsolescence of fixed silicon solutions.
The AMD-Xilinx's K26 SOM is an adaptive computing platform that enables the development of high-performance, production-ready AI systems at the edge in shorter development times \cite{k26somIdeal}.
This SOM features a XCK26-SK-KV260-G, which is a custom-built Zynq UltraScale+ MPSoC with a 64-bit quad-core ARM Cortex-A53 processor (1.333GHz of maximum theoretical frequency), a dual-core Cortex-R5F real-time processor (533GHz of maximum theoretical frequency) and an ARM Mali-400MP2 3D graphics processor in the Processing System (PS) connected to a 16nm FinFET Programmable Logic (PL) with access to a 4 GB 64-bit wide, 2400Mb/s DDR4 external memory \cite{k26somIdeal}.
The PL can host up to 4 DPU cores to accommodate different neural network architectures to be efficiently processed using high speed data pipes and parallel processing elements with fixed-point arithmetic units \cite{dpu}.

\subsection{Processor implementation}
The quantized lightweight FCN processor has been implemented embedding only one DPUCZDX8G IP core with B4096 architecture.
There is no point in using a parallel DPU configuration since this FCN model processes full images in a single pass with no previous image patching.
In this configuration, 4096 operations are performed per DPU clock cycle, which is was 300MHz.
More specifically, at each clock cycle 8 pixels of 16 channels of the input feature map are multiplied by the corresponding weights of 16 convolutional filters, thus 8x16x16 = 4096 multiplication operations and sums are performed.
Consequently, the theoretical raw compute power of this DPU is 1.229 TOPs.
The logic resources occupied by this particular implementation are 50,322 LUTs (42,97\%), 99,035 flip-flops (42,27\%), 75 BRAM (52,08\%), 48 dual-port URAMs (75,00\%), and 710 DSP48E2 slices (56,89\%).

\begin{figure}[t]
\centerline{\includegraphics[width=9cm]{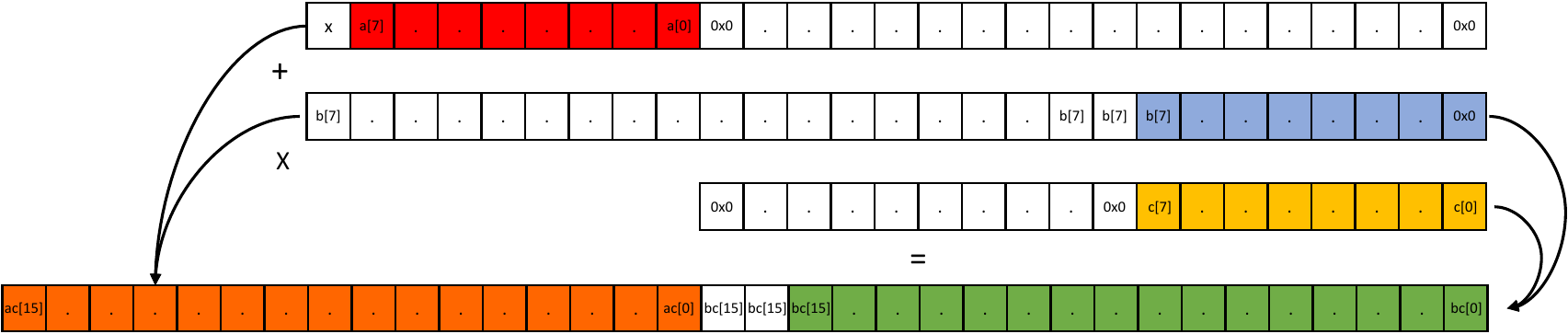}}
\caption{Packing two INT8 MAC with one DSP48E2 slice (adapted from \cite{fu2016deep}).}
\label{fig:packingTwoINT8}
\end{figure}

\begin{figure}[b]
\centerline{\includegraphics[width=7cm]{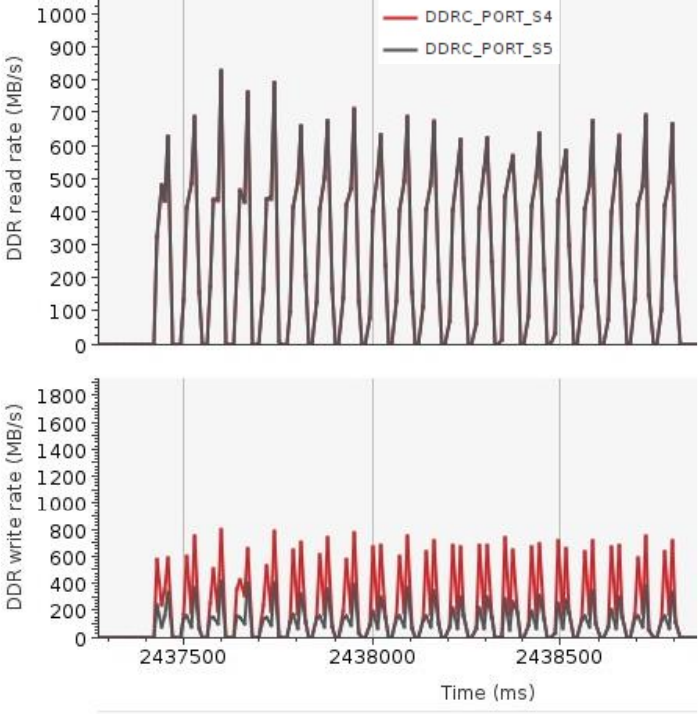}}
\caption{DDR data ports read/write rates as the DPU segments 20 images.}
\label{fig:DDRdataTransfer}
\end{figure}

The use of 27x18 multipliers enables the DPU to perform two concurrent INT8 multiplications increasing the overall throughput as shown in Fig. \ref{fig:packingTwoINT8}, although it is required to apply a specific technique to correctly accumulate output products and prevent overflow \cite{fu2016deep}.
On-chip memory, including BRAM and URAM, is used to store input feature-maps, intermediate activations, and output feature maps in order to improve throughput.
Data are reused as much as possible to reduce the accesses to external memory which, nevertheless, is necessary as the model (Table \ref{tab:compCompQuant}) does not fit in on-chip memory.
The average DDR memory access bandwith, which takes into account loading/storing of feature maps, weights and biases from/to DDR to/from DPU bank memory is 1953.778 MB/s (Fig. \ref{fig:DDRdataTransfer}).

\subsection{Experimental results and system validation}
The implemented quantized system underwent the same experimental testing used to validate the floating-point version.
Table \ref{tab:DPUmetrics} summarizes the average segmentation metrics across the 5 folds on the test subsets for both a 32-bit floating-point unquantized model and the 8-bit integer quantized model when executed on the CPU and DPU of the MPSoC respectively.
Figures in Table \ref{tab:DPUmetrics} demonstrate that the custom quantization process has resulted in minimal degradation of the segmentation metrics, with only a 0.18\% decrease in global IoU and 0.24\% decrease in weighted IoU.
The formulas used to calculate these metrics can be found in \cite{gutierrez22}

\begin{table}[t]
\caption{Average test segmentation metrics (Recall, Precision and IoU) of the floating-point model and the 8-bit integer models when run on the MPSoC CPU and DPU respectively.}
\label{tab:DPUmetrics}
\centering
\resizebox{8cm}{!}{
\begin{tabular}{|c|ccc|ccc|}
\hline
\textbf{Quantization} & \multicolumn{3}{c|}{\textbf{FP 32 (CPU)}} & \multicolumn{3}{c|}{\textbf{INT 8 (DPU)}} \\ \hline
 \multicolumn{1}{|c|}{\diagbox[]{\textbf{Class}}{\textbf{Metric}}} & \multicolumn{1}{c|}{\textbf{Rec.}} & \multicolumn{1}{c|}{\textbf{Prec.}} & \multicolumn{1}{c|}{\textbf{IoU}} & \multicolumn{1}{c|}{\textbf{Rec.}} & \multicolumn{1}{c|}{\textbf{Prec.}} & \multicolumn{1}{c|}{\textbf{IoU}}  \\ \hline
\textbf{Road} & \multicolumn{1}{c|}{98.63} & \multicolumn{1}{c|}{97.29} & \multicolumn{1}{c|}{96.12} & \multicolumn{1}{c|}{98.67} & \multicolumn{1}{c|}{97.20} & \multicolumn{1}{c|}{95.94} \\ \hline
\textbf{Road Marks} & \multicolumn{1}{c|}{86.70} & \multicolumn{1}{c|}{86.37} & \multicolumn{1}{c|}{76.78} & \multicolumn{1}{c|}{87.62} & \multicolumn{1}{c|}{85.26} & \multicolumn{1}{c|}{76.14} \\ \hline
\textbf{Vegetation} & \multicolumn{1}{c|}{94.22} & \multicolumn{1}{c|}{97.70} & \multicolumn{1}{c|}{93.32} & \multicolumn{1}{c|}{96.63} & \multicolumn{1}{c|}{96.40} & \multicolumn{1}{c|}{93.28} \\ \hline
\textbf{Sky} & \multicolumn{1}{c|}{84.97} & \multicolumn{1}{c|}{97.96} & \multicolumn{1}{c|}{87.09} & \multicolumn{1}{c|}{90.84} & \multicolumn{1}{c|}{96.49} & \multicolumn{1}{c|}{87.80} \\ \hline
\textbf{Others} & \multicolumn{1}{c|}{85.03} & \multicolumn{1}{c|}{80.89} & \multicolumn{1}{c|}{72.70} & \multicolumn{1}{c|}{80.22} & \multicolumn{1}{c|}{87.48} & \multicolumn{1}{c|}{71.93} \\ \hline
\textbf{Global} & \multicolumn{1}{c|}{95.34} & \multicolumn{1}{c|}{95.52} & \multicolumn{1}{c|}{92.28} & \multicolumn{1}{c|}{95.76} & \multicolumn{1}{c|}{95.72} & \multicolumn{1}{c|}{92.10} \\ \hline
\textbf{Weighted} & \multicolumn{1}{c|}{86.74} & \multicolumn{1}{c|}{89.80} & \multicolumn{1}{c|}{80.55} & \multicolumn{1}{c|}{88.17} & \multicolumn{1}{c|}{89.79} & \multicolumn{1}{c|}{80.31} \\ \hline
\end{tabular}}
\end{table}

\begin{table}[b]
\caption{Measured power consumption and required energy per image}
\label{power}
\centering
\resizebox{8cm}{!}{
\begin{tabular}{|c|ccc|}
\hline
\textbf{Framework (Accelerator)} & \multicolumn{1}{c|}{\textbf{TRT (GPU)}} & \multicolumn{1}{c|}{\textbf{TF Lite (CPU)}} & \multicolumn{1}{c|}{\textbf{Vitis AI (FPGA)}} \\ \hline
\textbf{Platform} & \multicolumn{1}{c|}{\textbf{Jetson Nano}} & \multicolumn{1}{c|}{\textbf{KV260}} & \multicolumn{1}{c|}{\textbf{KV260}} \\ \hline
\textbf{Power (W)} & \multicolumn{1}{c|}{6.020} & \multicolumn{1}{c|}{3.478} & \multicolumn{1}{c|}{7.635} \\ \hline
\textbf{Latency (s)} & \multicolumn{1}{c|}{0.110} & \multicolumn{1}{c|}{1.3479} & \multicolumn{1}{c|}{0.071} \\ \hline
\textbf{Energy (J)} & \multicolumn{1}{c|}{0.668} & \multicolumn{1}{c|}{4.688} & \multicolumn{1}{c|}{0.540}  \\ \hline
\end{tabular}}
\end{table}

\begin{figure}[h!]
\caption{Segmentation of the lightweight FCN on interurban scenarios. Rows: top, ground-truth; center, segmentation and bottom, false color. Columns: far left, image 617 (winter, cloudy, morning); left, image 209 (spring, rainy, morning); right, image 665 (winter, cloudy, afternoon) and far right, image 633 (autumn, cloudy, midday).}
\label{fig:dpuInterurban}
\vspace{0.1cm}
\centering
\includegraphics[width=4.4cm]{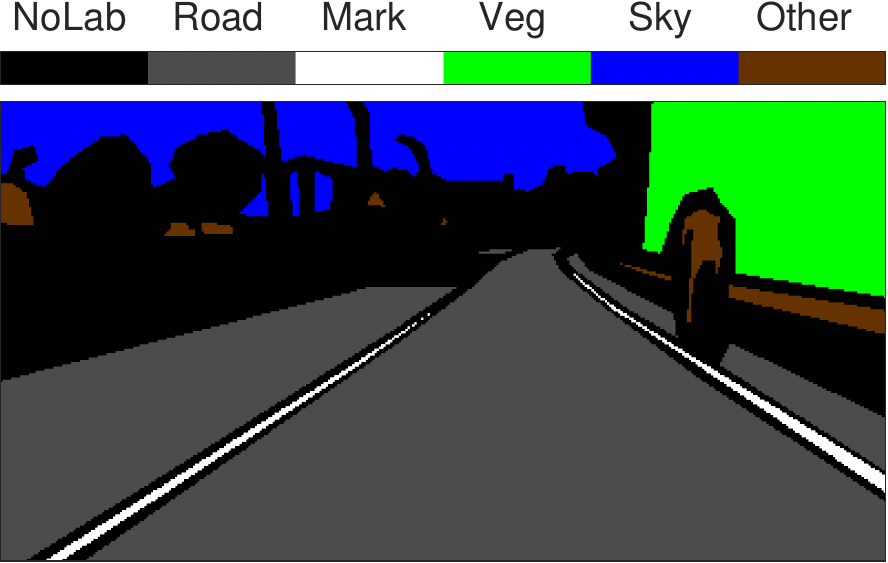}
\label{fig:2_617}
\includegraphics[width=4.4cm]{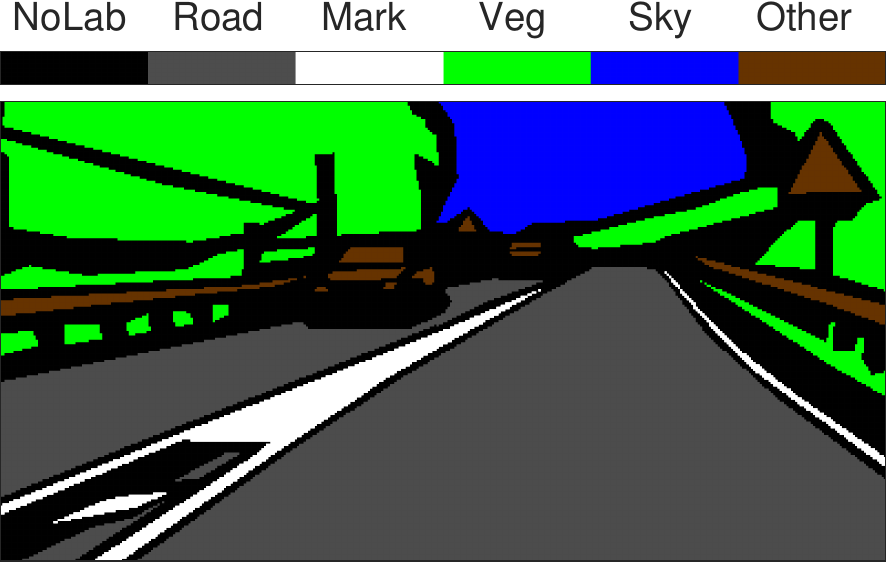}
\label{fig:2_209}
\includegraphics[width=4.4cm]{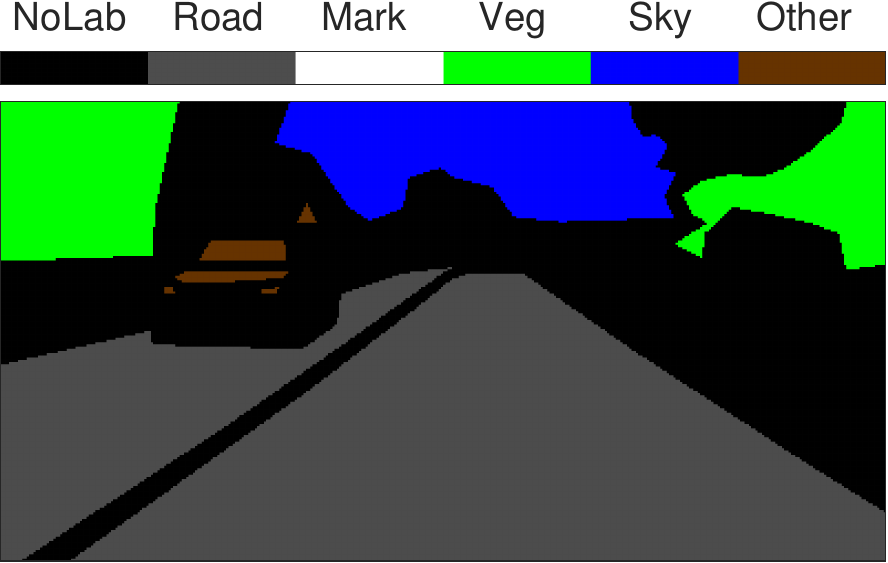}
\label{fig:2_665}
\includegraphics[width=4.4cm]{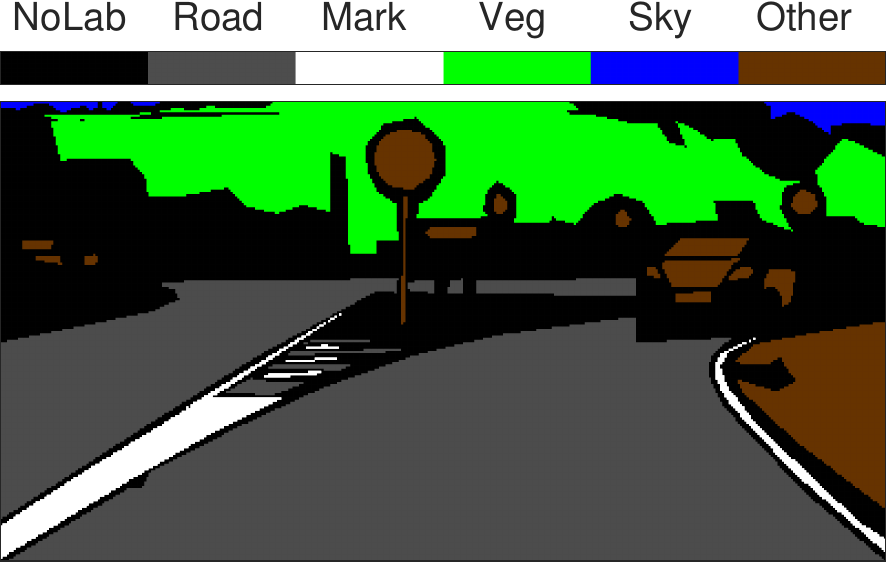}
\label{fig:2_633}\\
\vspace{0.1cm}
\includegraphics[width=4.4cm]{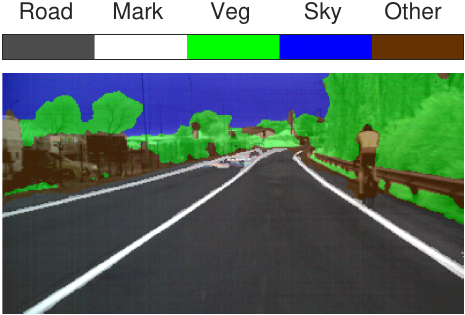}
\label{fig:1_617}
\includegraphics[width=4.4cm]{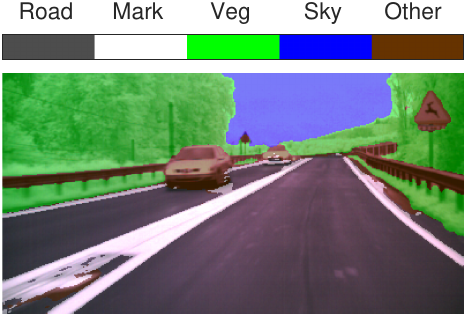}
\label{fig:1_209}
\includegraphics[width=4.4cm]{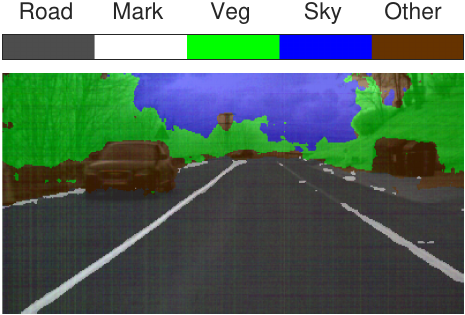}
\label{fig:1_665}
\includegraphics[width=4.4cm]{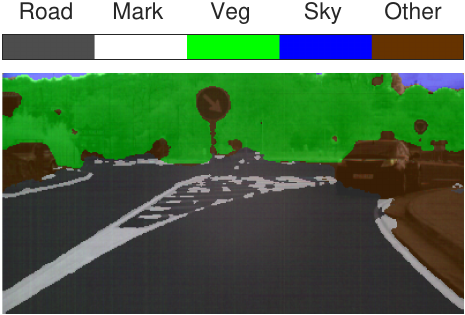}
\label{fig:1_633}\\
\vspace{0.1cm}
\includegraphics[width=4.4cm]{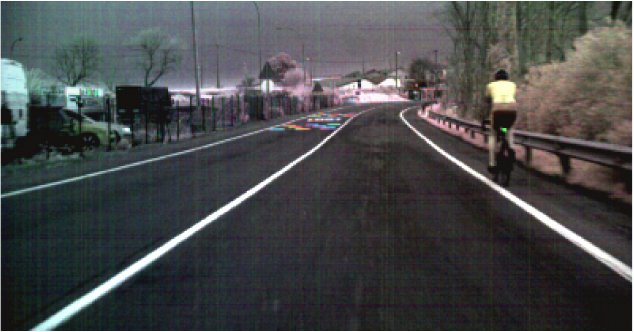}
\label{fig:3_617}
\includegraphics[width=4.4cm]{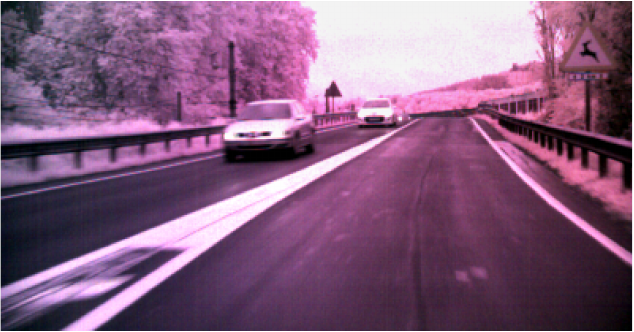}
\label{fig:3_209}
\includegraphics[width=4.4cm]{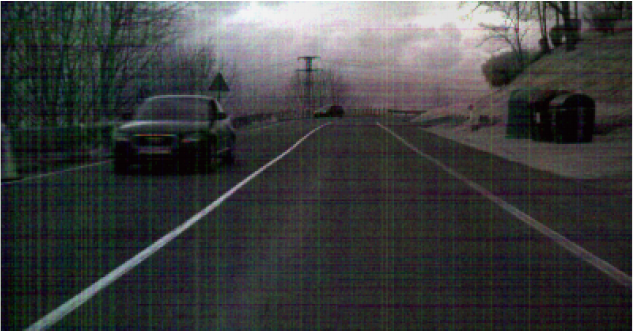}
\label{fig:3_665}
\includegraphics[width=4.4cm]{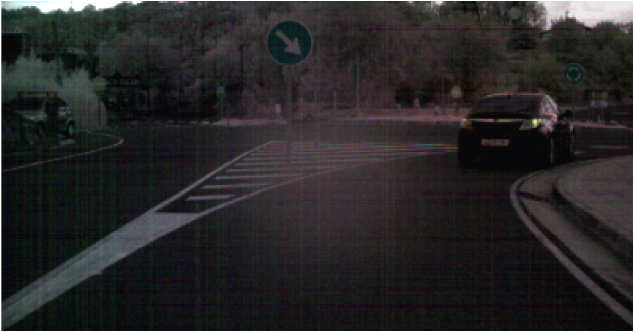}
\label{fig:3_633}
\end{figure}

\begin{figure}[h!]
\caption{Segmentation of the lightweight FCN on highway and urban scenarios. Rows: top, ground-truth; center, segmentation and bottom, false color. Columns: far left, image 637 (winter, sunny, midday, urban); left, image 626 (winter, cloudy, morning, urban); right, image 18 (summer, sunny, midday, highway) and far right, image 307 (winter, cloudy, midday, highway).}
\label{fig:dpuHighwayUrban}
\vspace{0.1cm}
\centering
\includegraphics[width=4.4cm]{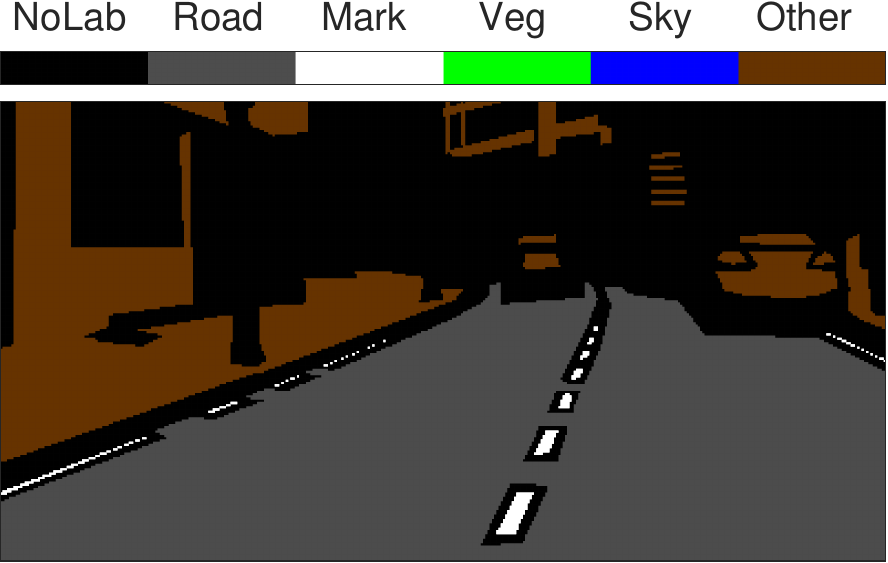}
\label{fig:2_637}
\includegraphics[width=4.4cm]{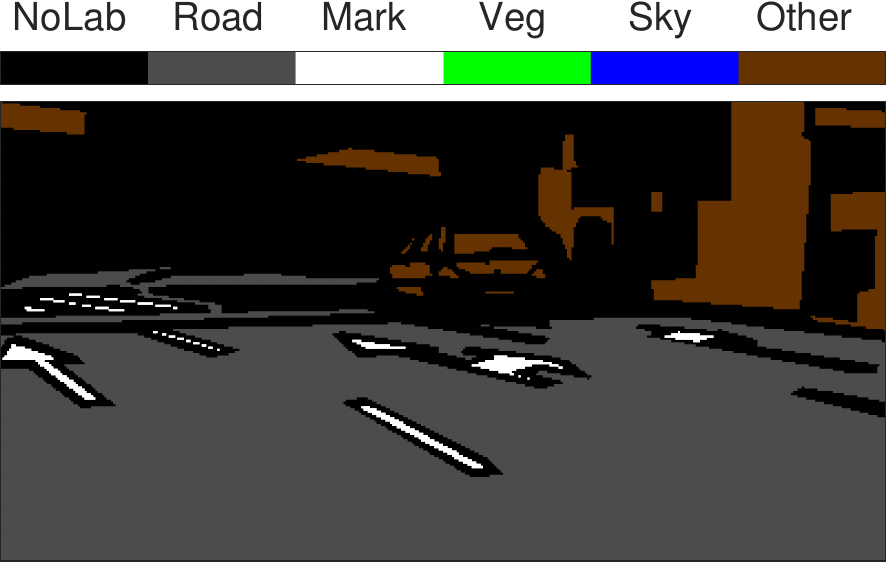}
\label{fig:2_626}
\includegraphics[width=4.4cm]{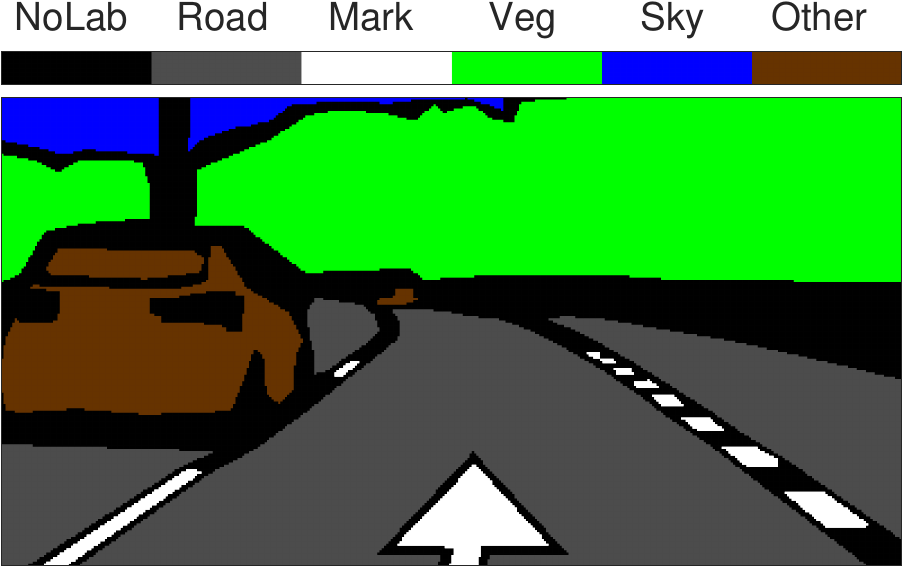}
\label{fig:2_018}
\includegraphics[width=4.4cm]{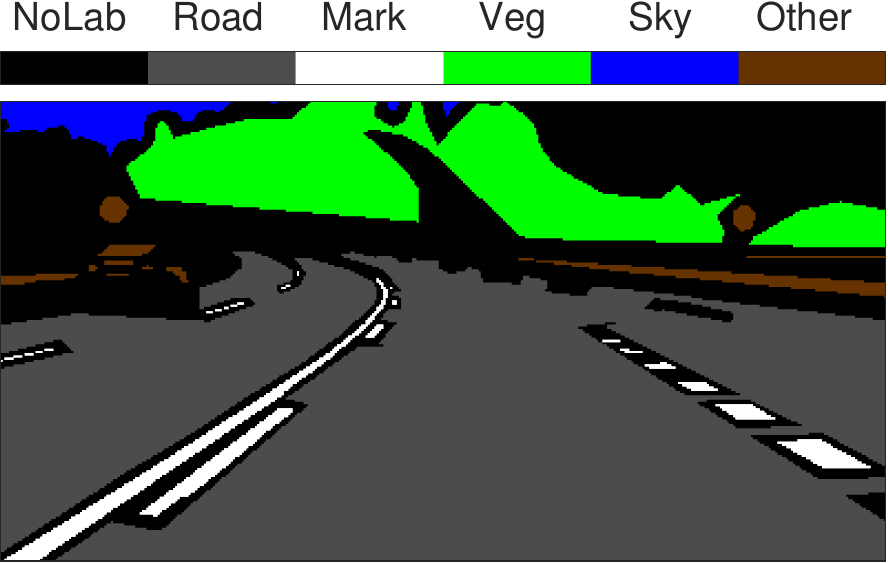}
\label{fig:2_307}\\
\vspace{0.1cm}
\includegraphics[width=4.4cm]{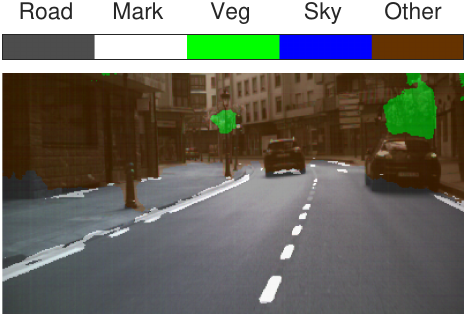}
\label{fig:1_637}
\includegraphics[width=4.4cm]{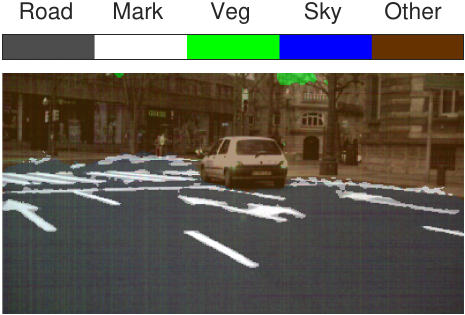}
\label{fig:1_626}
\includegraphics[width=4.4cm]{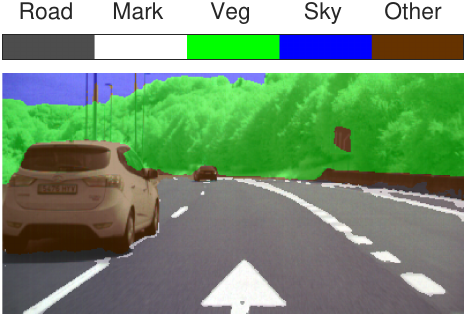}
\label{fig:1_018}
\includegraphics[width=4.4cm]{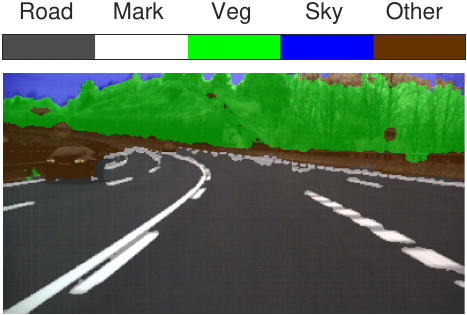}
\label{fig:1_307}\\
\vspace{0.1cm}
\includegraphics[width=4.4cm]{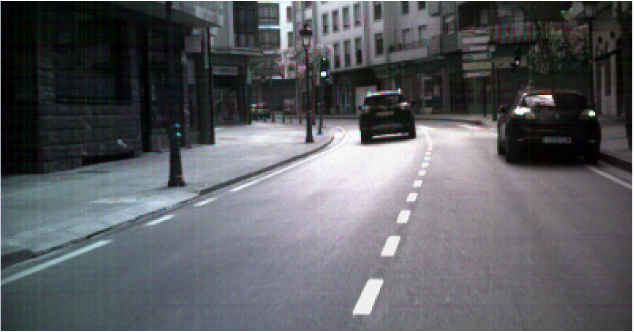}
\label{fig:3_637}
\includegraphics[width=4.4cm]{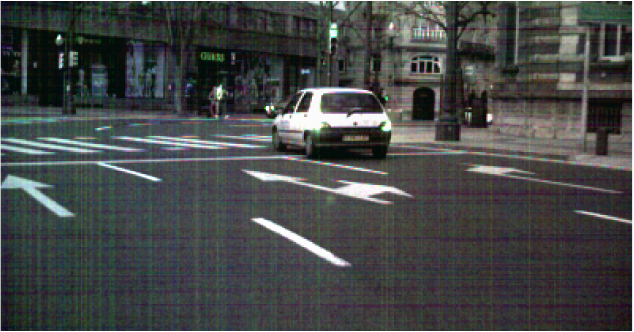}
\label{fig:3_626}
\includegraphics[width=4.4cm]{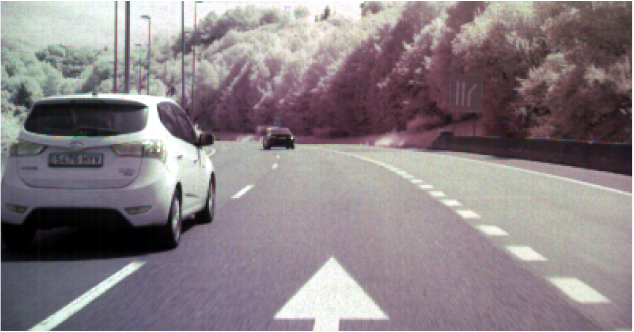}
\label{fig:3_018}
\includegraphics[width=4.4cm]{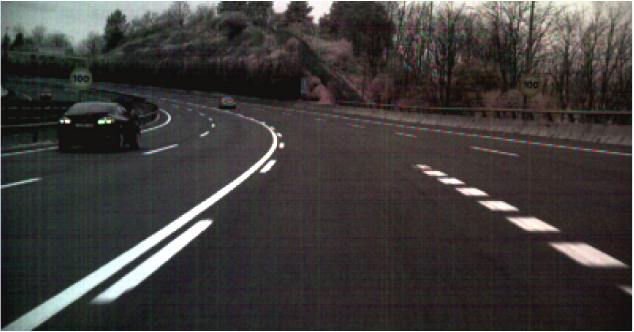}
\label{fig:3_307}
\end{figure}

Since the HSI-Drive dataset consists of weakly labeled data, it is convenient to complete the validation by visually evaluating the segmentation performance on complete images.
To illustrate this, we present in Fig. \ref{fig:dpuInterurban} and Fig. \ref{fig:dpuHighwayUrban} the inference results obtained on some representative scenes when the FCN is executed on the SOM.
Fig. \ref{fig:dpuInterurban} showcases the segmentation results on 4 interurban scenarios under different lighting and weather conditions.
The FCN demonstrates accurate detection and segmentation of objects requiring immediate attention, such as cars and a cyclist, as well as most of the information panels within a certain distance.
However, some segmentation errors occur in the background where the spatial resolution is low and spectral mixing is severe, or in particularly challenging situations, as shown in the image with a rain droplet on the lens (far right).
Fig. \ref{fig:dpuHighwayUrban} illustrates the segmentation results for 2 urban and 2 highway scenarios.
Urban images pose a significant challenge due to the presence of multiple overlapping objects with various sizes and shapes, as well as high lighting contrast caused by shadows.
Conversely, highway scenes are comparatively easier as the background is predominantly composed of sky or vegetation, which exhibit good separability indexes.
Overall, the FCN achieves satisfactory segmentation quality.

Regarding computational performance, it should be noted that this implementation uses only one software thread on the Quad-Core ARM-A53 since images are transmitted serially from the camera at a specified rate (11 FPS in our experimental setup) and there is no room for parallelism.
Experimentally measured processing throughput increases as the number of processed cubes increases, up to a maximum of 14.14 FPS.
For comparison purposes, we have also run the same lightweight FCN on a high-end XCZU7EV-2FFVC1156 device (ZCU104 evaluation board) by embedding 2 B4096 DPU cores.
Interestingly, the execution of 20 consecutive images with just 1 thread does not improve the performance obtained in the KV260, as the measured top value was 13.49 FPS.
Certainly, with this implementation, it is possible to reach a 40.39 FPS throughput processing 20 images simultaneously using 4 threads, but this parallelized configuration is not applicable to our case.
Summing up, deployment in the KV260 reaches 971.223 GOPs, with a theoretical peak performance of 1200 GOPs.
To provide context for the obtained FPS values, the three stages of the pipeline of the entire system have to be considered: image acquisition and communication, hyperspectral cube preprocessing, and image segmentation.
On the one hand, our camera setup records and transmits images at 11 FPS, while the preprocessing stage code, combining thread-level parallelism (OpenMP pragmas) with data-level parallelism (SIMD, via Neon), achieves a maximum processing rate of 14.61 FPS running on the Quad-Core ARM Cortex A53.
As a result, both the cube preprocessing and segmentation stages are able to keep up with the current frame rate of the camera setup.
Nevertheless, we aspire to improve these figures in the future.

Compared to other competing technologies, the power consumption of SRAM-based FPGAs is considered high.
However, this is compensated by the achievable computing performance if careful algorithm adaptation and processor design is carried out.
Power consumption monitoring on the KV260 SOM was performed using the \textit{platformstats} application.
This application communicates with the current sense device (INA260) through the I2C bus address 0x40 and displays the power consumption on the $VCC_{SOM}$ power rail.
The measured power consumption of the SOM was 2.350W when PL is powered down in idle state, 3.130W when the PL is powered up but not programmed, and 5.170W when the PL is powered up and programmed, but not running.
During the power consumption measurement of the segmentation system up and running on the SOM, which involved repeating the segmentation of 20 images 100 times, the average power consumption was 7.635W.
We compared the power consumption, image processing latency, and required total energy per image when implementing 8-bit quantized FCN models on a Jetson Nano embedded GPU-based SoC, a software version running on the Kria KV260's PS (PL is powered off), and on the Kria KV260's PL using the DPU.
Table \ref{tab:DPUmetrics} summarizes measured values.
As can be seen, the software execution on the CPU consumes the least power, but it is also the slowest, so the energy required to process each image is the highest.
The Jetson Nano SOM consumes about 1.5W less power than the KV260 SOM but is 1.5 times slower, so the FPGA-based SOM is still more energy-efficient for this application.

\section{Conclusions}
Small-size, snapshot-type hyperspectral cameras are enabling the use of HSI in novel application domains, including ADS.
The development of robust HSI processing systems on adaptive, low-cost hardware platforms with low latency and reduced power consumption is necessary for the gradual penetration of this technology in the field of autonomous driving.

As described in this article, achieving this goal involves both, a careful design of lightweight while capable neural network models, and the tailoring of the signal processing flows to the characteristics of domain-specific AI coprocessors.
This approach enables the development of an efficient system that effectively leverages hardware resources while ensuring that the segmentation quality does not deteriorate during the quantization process.
As described in this paper, the implementation process carried out on an MPSoC-based SOM has shown to be a successful approach to deploy practical HSI segmentation systems as it provides cost and development time savings while allowing for meeting the demanding performance requirements of ADAS/ADS.
The system has been tested and characterized in terms of memory footprint, latency, and power consumption including the cube preprocessing and the inference in the FCN.
The evaluation is performed using images obtained with a snapshot hyperspectral camera in real driving scenarios, encompassing diverse lighting and weather conditions.
This ensures that the obtained results are realistic and can be extrapolated to a potential real-world implementation based on the same technology.

%\onecolumn

%\begin{multicols}{2}
%    \small
%    \renewcommand{\refname}{ \vspace{-\baselineskip}\vspace{-1.1mm} }
%    \bibliographystyle{IEEEtran}
%    \bibliography{IEEEabrv, mybiblio}
%\end{multicols}

\bibliographystyle{IEEEtran}
\bibliography{IEEEabrv, mybiblio}

\onecolumn

\end{document}